\documentclass[letterpaper, 10 pt, conference]{ieeeconf}  

\IEEEoverridecommandlockouts                              

\usepackage{amsmath} 
\usepackage{amssymb}  

\usepackage{adjustbox}
\usepackage[ruled, linesnumbered]{algorithm2e}
\SetKwComment{Comment}{/* }{ */}
\SetAlFnt{\footnotesize}
\SetNlSty{normalsize}{}{.}
\DontPrintSemicolon
\usepackage{setspace}

\usepackage{cite}
\usepackage{cleveref}
\usepackage{wrapfig}
\usepackage{booktabs}
\usepackage{relsize}
\usepackage{multirow}
\usepackage{nicefrac}
\usepackage[dvipsnames]{xcolor}

\usepackage{amssymb}
\usepackage{pifont}
\newcommand{\cmark}{\ding{51}}%
\newcommand{\xmark}{\ding{55}}%
\title{\LARGE \bf
Voronoi-based Second-order Descriptor with\\Whitened Metric in LiDAR Place Recognition
\thanks{This work was partly supported by grants from the IITP (RS-2021-II211343-GSAI/10\%, RS-2022-II220951-LBA/15\%, RS-2022-II220953-PICA/15\%), the NRF (RS-2024-00353991-SPARC/15\%, RS-2023-00274280-HEI/15\%), the KEIT (RS-2025-25453780/15\%), and the KIAT (RS-2025-25460896/15\%), funded by the Korean government.}
}

\author{Jaein Kim$^{1}$, Hee Bin Yoo$^{2\ast}$, Dong-Sig Han$^{3\ast}$, and Byoung-Tak Zhang$^{1,4}$
\thanks{$^{1}$Interdisciplinary Program in Neuroscience, Seoul National University}
\thanks{$^{2}$Département d’Informatique, École Normale Supérieure (ENS)}
\thanks{$^{3}$Department of Computing, Imperial College London}
\thanks{$^{4}$Dept. of Computer Science and Engineering, Seoul National University}
\thanks{$^{\ast}$The majority of the work for this publication was done while these authors were in Seoul National University.}
}
\begin{document}

\maketitle

\thispagestyle{empty}
\pagestyle{empty}

\begin{abstract}

The pooling layer plays a vital role in aggregating local descriptors into the metrizable global descriptor in the LiDAR Place Recognition (LPR). In particular, the second-order pooling is capable of capturing higher-order interactions among local descriptors. However, its existing methods in the LPR adhere to conventional implementations and post-normalization, and incur the descriptor unsuitable for Euclidean distancing. Based on the recent interpretation that associates NetVLAD with the second-order statistics, we propose to integrate second-order pooling with the inductive bias from Voronoi cells. Our novel pooling method aggregates local descriptors to form the second-order matrix and whitens the global descriptor to implicitly measure the Mahalanobis distance while conserving the cluster property from Voronoi cells, addressing its numerical instability during learning with diverse techniques. We demonstrate its performance gains through the experiments conducted on the Oxford Robotcar and Wild-Places benchmarks and analyze the numerical effect of the proposed whitening algorithm.

\end{abstract}

\section{Introduction}
Place recognition is a problem for searching the nearest place to the query place from the memory of the autonomous system~\cite{lowry2015visual,garg2021your}.
It has gained its renown as one of the significant topics in the robotics field due to its correlation to loop closure detection and global localization in the SLAM or navigation systems~\cite{masone2021survey,zhang2024lidar}.
The research on place recognition is distinguishable concerning the sensory format of the observed place information.
Especially, LiDAR Place Recognition (LPR) utilizes three-dimensional LiDAR to capture geometric information~\cite{zhang2024lidar} and is known to be more robust to illuminative noise~\cite{zhang2024lidar,uy2018pointnetvlad}.

Deep learning recently has become a dominant approach in the LPR; it utilizes the expressiveness of neural networks and has surpassed traditional approaches~\cite{chen2023deep}.
It generally reframes the task as the retrieval problem and learns the compressed representation of places with the application of contrastive learning techniques~\cite{ruiz2018survey,masone2021survey,izquierdo2024optimal}.
This representation is referred to as a global descriptor,
which is obtainable by aggregating a set of local features while preserving local information to ensure the separability~\cite{rahman2024c3r,komorowski2021minkloc3d,komorowski2022minkloc3dv2}.

The most representative pooling methods in the LPR are NetVLAD~\cite{arandjelovic2016netvlad} and GeM~\cite{radenovic2018fine}, adopted from the Visual Place Recognition (VPR) studies.
These pooling methods are recognized as the first-order pooling in the field~\cite{boureau2010theoretical,carreira2012semantic},
which utilizes the first-order statistics, e.g., average or maximum, to compress the set of local descriptors.
Although many studies have demonstrated the effectiveness of first-order pooling methods in the LPR~\cite{uy2018pointnetvlad,komorowski2021minkloc3d,xu2021transloc3d,komorowski2022minkloc3dv2}, the first-order pooling has a limitation that it cannot capture the higher-order interactions between local features~\cite{vidanapathirana2021locus,rahman2024c3r}.

The second-order pooling utilizes second-order statistics between local features and embeds the pairwise correlation that cannot be modeled in first-order pooling~\cite{carreira2012semantic,lin2017improved,vidanapathirana2021locus,vidanapathirana2022logg3d}.
The limitation of previous methods is that they only exploit naive second-order statistics, e.g. the max or average operation over the outer products of local descriptors, and adhere to conventional normalization techniques irrelevant to the descriptor property.
However, \cite{qiu2024emvp} have recently suggested to interpret NetVLAD as the second-order pooling in contrast to the traditional perspective rooted in VLAD~\cite{jegou2010aggregating},
showing the equivalence between NetVLAD and the second-order operation with a condition on the soft-assignment.

\begin{figure}[t]
    \centering
    \includegraphics[width=\columnwidth]{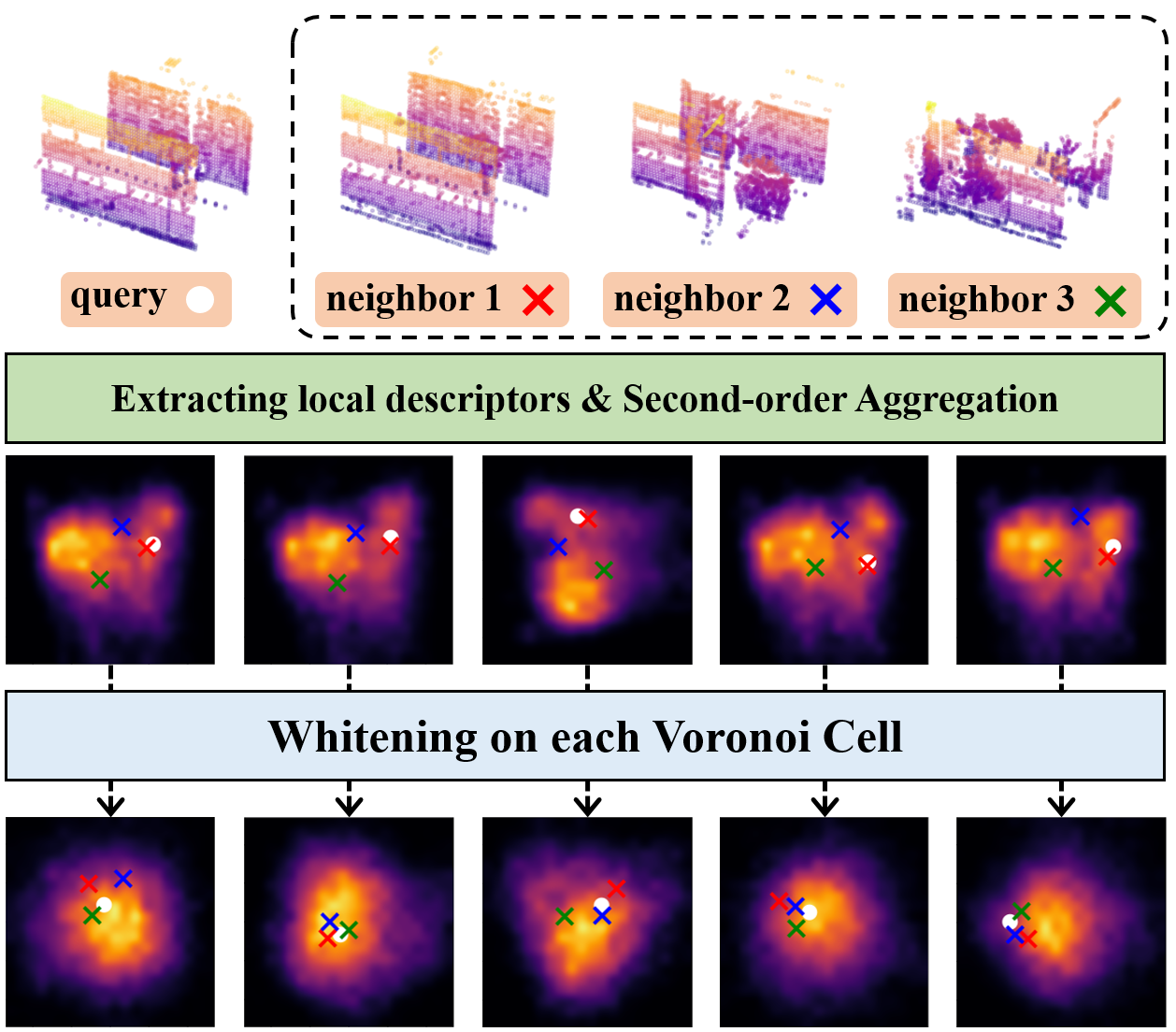}
    \caption{The illustration of feature space of each Voronoi cell learned by our method. We select the query place and its top-3 closest neighbors metrized by our method from the Oxford~\cite{maddern2017oxford} and visualize each cell of their descriptors reduced by ICA. White dots and colored cross marks denote query and neighbors. Our whitening transforms each Voronoi cell more homogeneous and suitable for Euclidean distancing.}\label{fig:vlad_metrics}
\end{figure}

The above interpretation expands the second-order pooling beyond a naive outer-product on local descriptors and provides an inductive bias to design more plausible normalization technique.
Our principal inductive bias from NetVLAD is that the clusters of global descriptor form independent subspaces with their own unique metrics,
in accordance with the fact that the clusters of VLAD are considered as embedded Voronoi cells~\cite{arandjelovic2013all,arandjelovic2016netvlad,chadha2017voronoi,lu2024supervlad}.
Despite such a property, existing methods using VLAD or second-order pooling naively apply $L^2$ normalization or linear maps to the global descriptor.
These might project the global descriptor onto a compact hypersphere or reduce the dimensionality, but they violate the topology of Voronoi cells and distort the metric of the descriptor.
We claim that a more relevant normalization exists for the descriptor space embedding Voronoi cells by relaxing the compactness condition.

This paper proposes a novel second-order pooling method applicable to a learning-based place recognition model.
It adopts the interpretation by \cite{qiu2024emvp} to justify the relation between NetVLAD and second-order pooling and designs simple neural networks to model the pooling modules.
From the understanding of the global descriptor as a set of disjoint clusters,
we propose to metrize the descriptor by the Mahalanobis distance considering the metric of each Voronoi cell,
which is achievable through whitening the feature of each Voronoi cell homogeneous using its covariance as \Cref{fig:vlad_metrics}.

Our core contribution is to realize the ZCA whitening~\cite{kessy2018whitening} with additional techniques to stabilize numerical errors, enabling the exact whitening on the learnable global descriptor in an end-to-end manner.
Our method demonstrates the state-of-the-art performance in Oxford RobotCar~\cite{maddern2017oxford} and Wild-Places~\cite{knights2023wildplaces} with the integration of second-order aggregation and suitable whitening to metrize the descriptor in Voronoi cells.
Numerical analyses also verify the positive effect of our whitening on the structure of each Voronoi cell.

\section{Related Works}
\paragraph{Learning-based LiDAR Place Recognition}
PointNetVLAD~\cite{uy2018pointnetvlad} and MinkLoc3D~\cite{komorowski2021minkloc3d} are some of the representative works in the LPR.
PointNetVLAD popularizes a general pipeline of learning-based retrieval models in the LPR with the utilization of NetVLAD~\cite{arandjelovic2016netvlad},
and MinkLoc3D introduces a sparse tensor convolution based on the MinkowskiEngine library~\cite{choy2019minkowski} and GeM pooling~\cite{radenovic2018fine}.
Most of the progress in the LPR has focused on proposing a novel backbone architecture~\cite{xu2021transloc3d,komorowski2022minkloc3dv2,xia2023casspr,qiu2024selfloc,shen2025forestlpr}
while adhering to the established first-order pooling methods.
As aforementioned, capturing more complex interactions is available through second-order statistics, and such approaches have shown its validity in the recent VPR studies~\cite{izquierdo2024optimal,qiu2024emvp}.

Second-order pooling, however, has not been studied as thoroughly as first-order pooling methods in the LPR.
Some representative works are Locus~\cite{vidanapathirana2021locus} and LoGG3D-Net~\cite{vidanapathirana2022logg3d} that aggregate local descriptors by O2P~\cite{carreira2012semantic} and apply the square root over the eigenvalues of the pooled descriptor.
Our approach extends prior methods by introducing a novel framework that establishes an association between second-order statistics and the cluster assumption.

\paragraph{VLAD and Descriptor Normalization}
Vector of Locally Aggregated Descriptors (VLAD)~\cite{jegou2010aggregating}, inspired by \textit{bag-of-features} and Fisher kernel~\cite{jaakkola1998exploiting}, accumulates the residuals of local descriptors to their centroids obtained by K-Means.
Some post-normalization techniques, e.g., signed square rooting~\cite{jegou2011aggregating,jegou2012negative} and intra-normalization~\cite{arandjelovic2013all},
have been proposed to compensate for VLAD by suppressing visual burstiness of overestimated features~\cite{jegou2009burstiness}.
However, their element-wise operations or $L^2$ normalization break the property of the VLAD descriptor as a set of clusters.

Meanwhile, a post-normalization by multivariate whitening has been studied previously; they whiten the global descriptor by PCA with a perspective coinciding with ours, viewing VLAD as the set of independent descriptors~\cite{jegou2012negative,chadha2017voronoi}.
However, they still normalize the global descriptor by $L^2$ at the last and are not suitable for learning neural networks in an end-to-end manner.
Our method relaxes the compactness condition and proposes to learn the metric with neural networks from scratch.

\section{Background: VLAD to Second-order Pooling}\label{sec:background}
This section introduces the derivation by \cite{qiu2024emvp} that associates NetVLAD~\cite{arandjelovic2016netvlad} with the second-order pooling,
based on its equivalence to the bilinear pooling~\cite{lin2015bilinear}.
Given the local descriptors projected on the Voronoi diagram $\mathbf{F}=\big[\mathbf{f}_1,\cdots,\mathbf{f}_{L}\big] \in \mathbb{R}^{C \times L}$,
the centroids of clusters $\mathbf{C}=\big[\mathbf{c}_1,\cdots,\mathbf{c}_{M}\big] \in \mathbb{R}^{C \times M}$,
and the soft assignment of each local descriptor to clusters $\mathbf{P}=\big[\mathbf{p}_1,\cdots,\mathbf{p}_{L}\big] \in \mathbb{R}^{M \times L}$,
the global descriptor $\widetilde{\mathbf{X}}$ by NetVLAD is formulated as follows:
\begin{equation}\label{eq:bilinear_netvlad}
    \widetilde{\mathbf{X}} = \sum^{L}_{i=1} \big[ \mathbf{f}_i-\mathbf{c}_1, \cdots,  \mathbf{f}_i-\mathbf{c}_M\big]\;\odot\;{\underbrace{\big[ \mathbf{p}_{i},\cdots,\mathbf{p}_{i}\big]}_{C}}^{\top},
\end{equation}
where $\odot$ is an element-wise product.
NetVLAD computes $\mathbf{X}$ and $\mathbf{P}$ from the input local descriptors and parametrizes $\mathbf{C}$ as learnable weights.
\Cref{eq:bilinear_netvlad} is reducible further when the additional constraint on $\mathbf{P}$ is given,
saying $\sum^{L}_{i=1}\mathbf{p}_i=\mathbf{v}$ \textit{is a constant vector}. Then,
\begin{equation}\label{eq:reduced_netvlad}
\begin{aligned}
    (\ref{eq:bilinear_netvlad}) &= \mathbf{F}\,\mathbf{P}^{\top}-\mathbf{C}\;\odot\;\sum^{L}_{i=1}{\big[ \mathbf{p}_{i},\cdots,\mathbf{p}_{i}\big]}^{\top}\\
    &= \mathbf{F}\,\mathbf{P}^{\top} - \,\mathbf{C}\;\odot\;{\big[ \mathbf{v},\cdots,\mathbf{v}\big]}^{\top},
\end{aligned}
\end{equation}
where the cluster centroids $\mathbf{C}$ become independent to the input local descriptors.
We can declare that the global descriptor is pooled by second-order statistics and interpretable as clusters in Voronoi cells simultaneously, thanks to the disregardable centroids: $\widetilde{\mathbf{X}} \triangleq \mathbf{F}\,\mathbf{P}^{\top}$.
However, \cite{qiu2024emvp} remain to normalize the descriptor with heuristic approach based on an approximated matrix power normalization and intra-normalization.
We propose a suitable normalization technique that reflects the metrics of each Voronoi cell.

\section{Methods}
\begin{figure*}[t!]
    \centering
    \includegraphics[width=0.98\textwidth, trim={0.15cm 0.3cm 0.15cm 0.2cm}, clip]{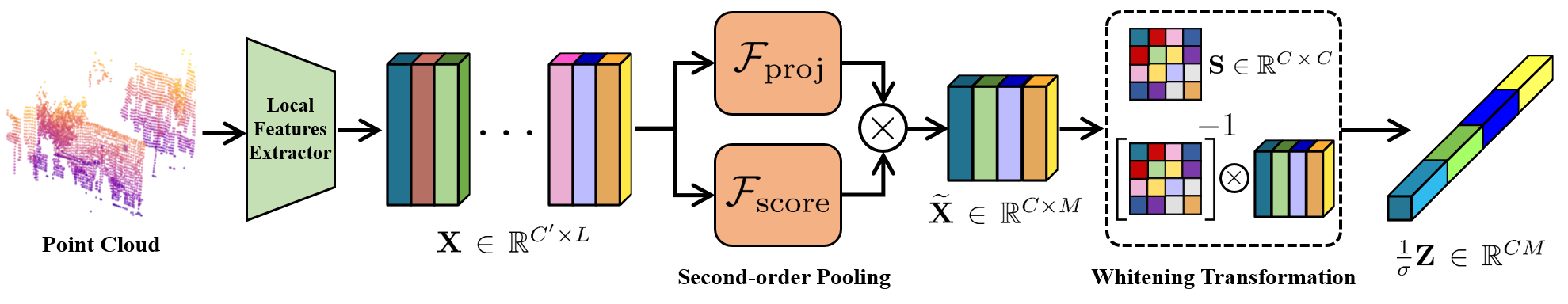}
    \caption{The overall architecture. First, the input local descriptors $\mathbf{X}$ are mapped by networks $\mathcal{F}_\text{proj}$ and $\mathcal{F}_\text{score}$, multiplied into the global descriptor $\widetilde{\mathbf{X}}$. Our whitening module implemented with ZCA whitening~\cite{kessy2018whitening} computes the standard deviation matrix $\mathbf{S}$ from $\widetilde{\mathbf{X}}$ and normalizes each cluster's feature by multiplying $\mathbf{S}^{-1}$ over $\widetilde{\mathbf{X}}$. Finally, the global descriptor $\mathbf{Z}$ is vectorized and scaled down to $\frac{1}{\sigma}\mathbf{Z}$.}
    \label{fig:overall_arch}
\end{figure*}
Our pooling method consists of the feature projection $\mathcal{F}_\text{proj}(\cdot)$ and soft-assignment score $\mathcal{F}_\text{score}(\cdot)$ networks that correspond to $\mathbf{F}$ and $\mathbf{P}$ respectively,
and the whitening module.
As shown in \Cref{fig:overall_arch}, the local descriptors $\mathbf{X}$ are mapped into the compressed features and the assignment scores to each Voronoi cell.
Then the global descriptor $\widetilde{\mathbf{X}}$ is acquired from the summation of the projected features weighted by the score.
Finally, each cluster feature of the global descriptor is whitened, and the whole descriptor vector is scaled by a positive coefficient $\sigma$.

\subsection{The Second-order Descriptor in Voronoi Diagram and its Mahalanobis Distancing}
Let us consider that $\mathcal{F}_\text{proj}(\cdot)$ and $\mathcal{F}_\text{score}(\cdot)$ are neural networks with two linear layers whose hidden layers are followed by batch normalization~\cite{ioffe2015batch} and GELU~\cite{hendrycks2016gaussian} activation.
Given $\mathbf{X}=\big[\mathbf{x}_{1},\cdots,\mathbf{x}_{L}\big]$, which is the sequence of local descriptors of length $L$,
these networks map each local descriptor in $\mathbf{X}$ with shared weights and output $\mathcal{F}_\text{proj}(\mathbf{X})$ and $\mathcal{F}_\text{score}(\mathbf{X})$. 
Then the weighted summation of projected features in $\mathcal{F}_\text{proj}(\mathbf{X})$ by the scores toward each cluster $\mathcal{F}_\text{score}(\mathbf{X})$ is computed as \Cref{eq:multiplying}:
\begin{equation}\label{eq:multiplying}
\begin{gathered}
    \widetilde{\mathbf{X}} = \mathcal{F}_\text{proj}(\mathbf{X})\;\text{softmax}(\mathcal{F}_\text{score}(\mathbf{X}))^\top,\\\text{s.t.}\;\mathcal{F}_\text{proj}(\mathbf{X})\in\mathbb{R}^{C \times L},\;\mathcal{F}_\text{score}(\mathbf{X})\in\mathbb{R}^{M \times L}, 
\end{gathered}
\end{equation}
where $\mathbf{\widetilde{X}}$ is the aggregated global descriptor, and $\text{softmax}(\cdot)$ is applied along the sequence dimension of $\mathcal{F}_\text{score}(\mathbf{X})$ to meet the constraint mentioned in \Cref{sec:background}, following \cite{qiu2024emvp}.

The global descriptor $\widetilde{\mathbf{X}}$ is interpretable as the set of features in the Voronoi diagram.
Since the Voronoi cells are independent of each other, let us assume that the distribution of each cell follows Gaussian with its own unique covariance $\mathbf{\Sigma}_i \in \mathbb{R}^{C \times C}$, where $i$ is an index of the cell.
The distribution of the entire descriptor space is explainable as the joint distribution of independent cells.
Hence, we suggest to distance global descriptors $\widetilde{\mathbf{X}}_1$ and $\widetilde{\mathbf{X}}_2$ by Mahalanobis distance, with the abuse of notation that omits the vectorization:
\begin{equation}\label{eq:mahalanobis}
\begin{gathered}
    d^{2}_\text{Mahal}(\widetilde{\mathbf{X}}_1, \widetilde{\mathbf{X}}_2;\mathlarger{\mathbf{\Sigma}})=
    (\widetilde{\mathbf{X}}_1-\widetilde{\mathbf{X}}_2)^\top\,\mathlarger{\mathbf{\Sigma}}^{-1}\;(\widetilde{\mathbf{X}}_1-\widetilde{\mathbf{X}}_2)\\
    =\sum^M_{i=1}\;(\widetilde{\mathbf{x}}_{1i}-\widetilde{\mathbf{x}}_{2i})^\top\,\mathlarger{\mathbf{\Sigma}}^{-1}_i\;(\widetilde{\mathbf{x}}_{1i}-\widetilde{\mathbf{x}}_{2i}),\\
    \text{s.t.}\;\;
    \widetilde{\mathbf{X}}_j = \big[\widetilde{\mathbf{x}}_{j1},\,\cdots,\;\widetilde{\mathbf{x}}_{jM}\big],\;\;
    \mathlarger{\mathbf{\Sigma}} =
    \begin{bmatrix}
        \mathlarger{\mathbf{\Sigma}}_1&\!\cdots&\!\mathbf{0}\\
        \vdots&\!\ddots&\!\vdots\\
        \mathbf{0}&\!\cdots&\!\mathlarger{\mathbf{\Sigma}}_M
    \end{bmatrix},
\end{gathered}
\end{equation}
where the covariance of joint distribution $\mathlarger{\mathbf{\Sigma}}$ is a block diagonal matrix.
Due to the independence between the Voronoi cells, the entire operation is decomposable into the summation of Mahalanobis distance inside each cell with the reduced cost compared to the operation by the full $\mathlarger{\mathbf{\Sigma}}$.

\subsection{Whitening Transformation for Mahalanobis Distancing}
We implement \Cref{eq:mahalanobis} with the whitening transformation instead of explicitly computing the inverse of covariances.
It efficiently converts the Mahalanobis distance into the Euclidean distance:
\begin{equation}\label{eq:naive_whitening}
\begin{gathered}
    d^{2}_\text{Mahal}(\widetilde{\mathbf{X}}_1, \widetilde{\mathbf{X}}_2;\mathlarger{\mathbf{\Sigma}})
    =\sum^{M}_{i=1}\;\| \mathbf{z}_{1i}-\mathbf{z}_{2i} \|^2_2\;,\\\text{s.t.}\;\;\mathbf{S}_i\mathbf{S}^{\top}_i=\mathlarger{\mathbf{\Sigma}}_{i}\,,\;\;\mathbf{z}_{ji} = \mathbf{S}_{i}^{-1}\mathbf{x}_{ji}.
\end{gathered}
\end{equation}

At a glance, one can acquire covariance $\mathlarger{\mathbf{\Sigma}}_i$ with the mini-batch samples for each cell;
however, the statistics acquired from the batch-wise samples during training may not reflect the domain shift during the evaluation~\cite{li2016revisiting,choi2021meta}.
As we measure the descriptor metric by Mahalanobis distance, the batch-wise covariance at the final layer can distort the metric during the evaluation.
Therefore, we propose to modify the whitening with the assumption that \textit{every cluster cell has the identical distribution parameters} as follows:
\begin{equation}\label{eq:instance_whitening}
\begin{gathered}
    d^{2}_\text{Mahal}(\widetilde{\mathbf{X}}_1, \widetilde{\mathbf{X}}_2)
    =\sum^{M}_{i=1}\;\| \mathbf{z}_{1i}-\mathbf{z}_{2i} \|^2_2\;,\\ \text{s.t.}\;\;\mathbf{z}_{ji} = \mathbf{S}_{j}^{-1}(\mathbf{x}_{ji}-\mathbf{\hat{\mu}}_{j})\,,\;\;\mathbf{\hat{\mu}}_{j}=\frac{1}{M}\sum^{M}_{i=1}{\mathbf{x}_{ji}}\\\mathbf{S}_{j}\mathbf{S}_{j}^{\top}=\frac{1}{M}\sum^{M}_{i=1}{(\mathbf{x}_{ji}-\mathbf{\hat{\mu}}_{j})(\mathbf{x}_{ji}-\mathbf{\hat{\mu}}_{j})^{\top}}.
\end{gathered}
\end{equation}
Samples for the estimates are taken \textit{within the instance} thanks to the assumption,
free from tracking the learnable distribution parameters during the training.

\begin{table*}[ht]
    \centering
    \caption{The average recall at top-1 and top-1\% retrievals in the Oxford~\cite{maddern2017oxford}. \textdagger: the results quoted from the original paper.}

    \label{tab:oxford}
    \resizebox{0.97\textwidth}{!}{{\tiny
    \begin{tabular}{l|cccccccccc}
        \toprule[0.4pt]
        \multirow{2}{*}{Methods} & \multicolumn{2}{c}{Oxford} & \multicolumn{2}{c}{U.S.} & \multicolumn{2}{c}{R.A.} & \multicolumn{2}{c}{B.D} & \multicolumn{2}{c}{Average}\\
        &&&&&&&&&&\\[-1.0em]
        & R@1 & R@1\% & R@1 & R@1\% & R@1 & R@1\% & R@1 & R@1\% & R@1 & R@1\% \\
        \hline
        &&&&&&&&&&\\[-1.0em]
        PointNetVLAD~\cite{uy2018pointnetvlad}       & 74.88 & 87.89 & 68.09 & 81.37 & 64.05 & 74.66 & 63.42 & 70.63 & 67.61 & 78.64\\
        &&&&&&&&&&\\[-1.0em]
        TransLoc3D~\cite{xu2021transloc3d} & 93.99 & 98.13 & 83.50 & 93.23 & 80.11 & 90.39 & 80.01 & 87.05 & 84.40 & 92.20\\
        &&&&&&&&&&\\[-1.0em]
        MinkLoc3D~\cite{komorowski2021minkloc3d}     & 93.76 & 97.91 & 86.01 & 95.04 & 81.11 & 91.19 & 82.66 & 88.48 & 85.89 & 93.16 \\
        &&&&&&&&&&\\[-1.0em]
        MinkLoc3Dv2~\cite{komorowski2022minkloc3dv2} & 96.26 & \underline{98.87} & 90.85 & 96.65 & 86.49 & 93.75 & 86.26 & 91.15 & 89.97 & 95.11 \\
        &&&&&&&&&&\\[-1.0em]
        CASSPR~\cite{xia2023casspr} & 95.84 & 98.75 & 92.91 & 98.00 & \underline{89.44} & 94.69 & 87.25 & \underline{92.36} & 91.36 & 95.95\\
        &&&&&&&&&&\\[-1.0em]
        SelFLoc~\cite{qiu2024selfloc}\textsuperscript{\textdagger} & 96.0 & 98.8 & \underline{93.2} & \underline{98.3} & 88.8 & \underline{94.8} & \underline{88.4} & 92.4 & \underline{91.6} & \underline{96.1} \\
        &&&&&&&&&&\\[-1.0em]
        Ours (C=16, M=16) & \underline{96.63} & \underline{98.87} & 91.04 & 97.36 & 86.95 & 93.89 & 87.55 & 91.76 & 90.54 & 95.47 \\ 
        &&&&&&&&&&\\[-1.0em]
        Ours (C=128, M=64) & \textbf{97.97} & \textbf{99.40} & \textbf{94.97} & \textbf{98.58} & \textbf{90.99} & \textbf{95.90} & \textbf{90.98} & \textbf{94.33} & \textbf{93.73} & \textbf{97.05} \\ 
        \bottomrule[0.4pt] 
    \end{tabular}
    }}
\end{table*}

\begin{algorithm}[t]
\setstretch{1.4}
\SetAlCapHSkip{0.5em}
\KwIn{$\widetilde{\mathbf{X}} = \big[\widetilde{\mathbf{x}}_1, \cdots,\widetilde{\mathbf{x}}_M\big] \in \mathbb{R}^{C\,\times M}$, a small noise $\varepsilon=1\mathrm{e}\!-\!5$}
\KwOut{$\mathbf{Z} = \big[\mathbf{z}_1, \cdots,\mathbf{z}_M\big] \in \mathbb{R}^{C\,\times M}$}
$\overline{\mathbf{X}} = \big[\widetilde{\mathbf{x}}_1-\mathbf{\hat\mu}, \cdots,\widetilde{\mathbf{x}}_M-\mathbf{\hat\mu}\big]$ from $\mathbf{\hat\mu} = \frac{1}{M}\sum^{M}_{i=1}\widetilde{\mathbf{x}}_i$\;
Initialize $\mathbf{\hat{\Sigma}}=\frac{1}{M}\,\overline{\mathbf{X}}\;\overline{\mathbf{X}}^{\top}$ and $\;\mathbf{\hat{F}}=\frac{\mathrm{Tr}(\mathbf{\hat{\Sigma}})}{C}\;\mathbf{I}$\;
$\rho_\text{RBLW}=\min\bigg( \dfrac{\big(\frac{M-2}{M}\big)\mathrm{Tr}(\mathbf{\hat{\Sigma}}^2) + \mathrm{Tr}^2(\mathbf{\hat{\Sigma}})}{(M+2)\big( \mathrm{Tr}(\mathbf{\hat{\Sigma}}^2) - \frac{1}{C}\mathrm{Tr}^2(\mathbf{\hat{\Sigma}}) \big)},\;1\bigg)$\;
$\mathbf{\hat{\Sigma}}_\text{RBLW}=\rho_\text{RBLW}\,\mathbf{\hat{F}}+(1-\rho_\text{RBLW})\,\mathbf{\hat{\Sigma}}$\;
$\mathbf{\hat{Q}}\mathbf{\hat{\Lambda}}\mathbf{\hat{Q}}^{\top}=\texttt{SVDPI}\big(\mathbf{\hat{\Sigma}}_\text{RBLW} + \varepsilon\,\mathbf{I}\big)$\;
$\mathbf{Z} = \mathbf{\hat{Q}}\mathbf{\hat{\Lambda}}^{-\frac{1}{2}}\mathbf{\hat{Q}}^{\top}\overline{\mathbf{X}}$\Comment*[r]{ZCA Whitening}
\Return $\mathbf{Z}$
\caption{The proposed ZCA whitening.}\label{alg:zca_whitening}
\end{algorithm}

\Cref{eq:instance_whitening} is realizable with the estimation of the distribution parameters among Voronoi cells and the application of ZCA whitening over each cell's feature.
However, a computation of the standard deviation matrix from a naive sample covariance results in the rank deficiency since the number of clusters is smaller than the dimension of cluster features,
providing only the limited amount of information.

We resolve this issue by implementing ZCA whitening with the Rao-Blackwell Ledoit-Wolf (RBLW)~\cite{chen2010shrinkage} algorithm for the covariance shrinkage as \Cref{alg:zca_whitening}.
The RBLW algorithm estimates the coefficient $\rho_\text{RBLW}$ that minimizes the difference between the true covariance, which is supposed to be full rank, and the combination of $\mathbf{\hat{F}}$ and $\mathbf{\hat{\Sigma}}$.

However, our pooling method, which applies ZCA whitening over the descriptor learned with neural networks in an end-to-end manner, is vulnerable to the gradient instability.
It is known that a naive eigendecomposition causes a gradient explosion when multiple eigenvalues have close values~\cite{ionescu2015matrix,ionescu2015training,wang2019backpropagation}.
Thus, we replace it with an algorithm stable regarding the backpropagation~\cite{wang2019backpropagation} (\texttt{SVDPI} in \Cref{alg:zca_whitening}),
which decomposes the covariance using SVD without auto differentiation and backpropagates the gradient using the iterative method.
Please refer to its original paper~\cite{wang2019backpropagation} for the details.

\subsection{The Configuration of Place Recognition Model}
We apply our pooling method to the backbone network from MinkLoc3Dv2~\cite{komorowski2022minkloc3dv2} as a default model.
The number of clusters $M$ and the dimension of cluster features $C$ are configured differently with regard to the size of global descriptor and datasets.
We also divide the whitened descriptor $\mathbf{Z}$ by a scalar hyperparameter $\sigma$ to normalize the norm of the entire descriptor.
Since the divided descriptor follows the Gaussian distribution with diagonal covariance, $\frac{1}{\sigma}\mathbf{Z}\!\sim\!\mathcal{N}(\mathbf{0}, \frac{1}{\sigma^2}\mathbf{I})$,
its Euclidean distancing still implies the the Mahalanobis.
The final metric is optimized by the Truncated Smooth-AP loss with the identical setups to \cite{komorowski2022minkloc3dv2},
but we substitute a learning rate scheduler with the cosine annealing scheduler~\cite{loshchilov2016sgdr}.

\section{Experimental Results}\label{sec:result}
We conducted the experiments with two place recognition benchmarks:
Oxford~\cite{maddern2017oxford} and Wild-Places~\cite{knights2023wildplaces} concerning the urban and natural environments.
Following the training and evaluation protocol from \cite{uy2018pointnetvlad},
the evaluations were also conducted with the In-house datasets which are out-of-distribution from the Oxford.
In the Wild-Places, we followed the protocol from \cite{knights2023wildplaces} and quantitatively assessed models with respect to the loop-closure detection and the global localization.
Ablation studies and analyses were also conducted to verify the backbone-agnostic applicability and the validity of numerical techniques in the ZCA whitening.

\subsection{Experiments on Oxford RobotCar}
\begin{figure}[t!]
    \centering
    \includegraphics[width=\columnwidth]{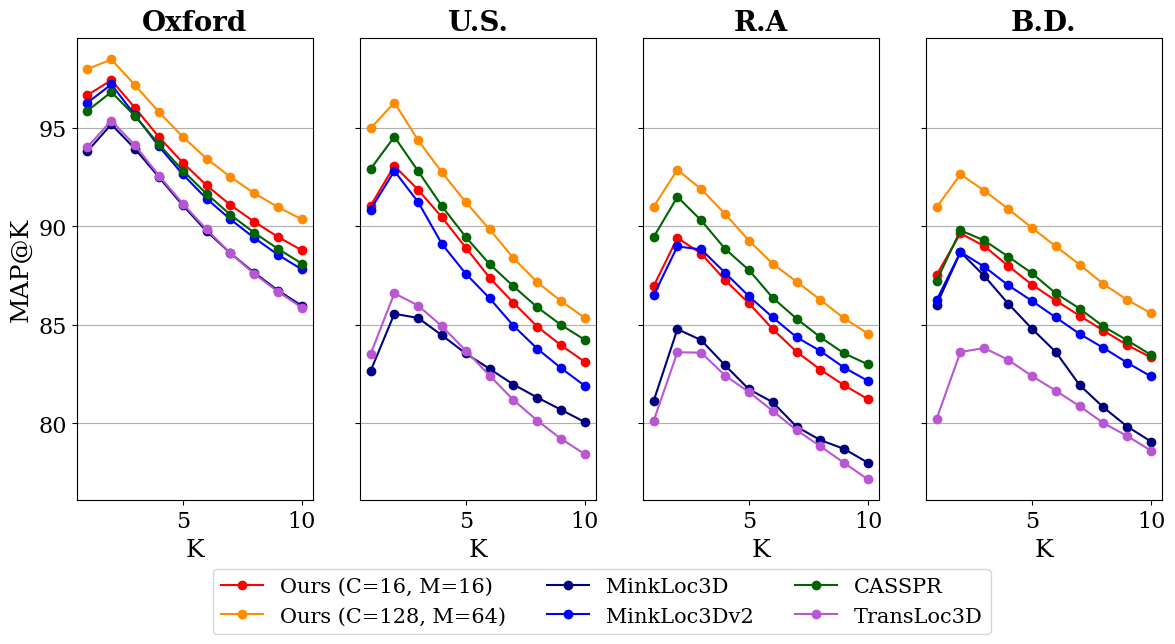}
    \caption{Mean Average Precision at K (MAP@K) plotted up to 10 predictions in the Oxford and In-house datasets.}
    \label{fig:curve_oxford}
\end{figure}

\begin{table*}[t!]
    \centering
    \caption{The intra-sequence and inter-sequence results in the Wild-Places~\cite{knights2023wildplaces}. \textdagger:~the results quoted from the original paper.}
    \resizebox{0.97\textwidth}{!}{{\scriptsize
    \begin{tabular}{l|c|cccccccc|cccc}
        \toprule[0.5pt]
        \multirow{3}{*}{Methods} & \multirow{3}{*}{\begin{tabular}[x]{@{}c@{}}Descriptor\\Dimensions\end{tabular}} & \multicolumn{8}{c|}{\textbf{Intra-sequence}} & \multicolumn{4}{c}{\textbf{Inter-sequence}} \\
        &&&&&&&&&&&&&\\[-1.0em]
        & & \multicolumn{2}{c}{V-03} & \multicolumn{2}{c}{V-04} & \multicolumn{2}{c}{K-03} & \multicolumn{2}{c|}{K-04} & \multicolumn{2}{c}{Venman} & \multicolumn{2}{c}{Karawatha}\\
        &&&&&&&&&&&&&\\[-1.0em]
        & & F1 & R@1 & F1 & R@1 & F1 & R@1 & F1 & R@1 & R@1 & MRR & R@1 & MRR \\
        \hline
        &&&&&&&&&&&&&\\[-0.8em]
        ScanContext~\cite{kim2018scan} & 1200 & 05.54 & 12.79 & 39.28 & 43.33 & 31.33 & 43.55 & 58.09 & 61.72 & 43.15 & 45.23 & 52.79 & 56.40 \\
        &&&&&&&&&&&&&\\[-0.8em]
        TransLoc3D~\cite{xu2021transloc3d} & 256 & 22.92 & 35.50 & 75.00 & 64.03 & 43.50 & 38.95 & 75.80 & 75.53 & 62.85 & 74.92 & 54.32 & 69.08 \\
        &&&&&&&&&&&&&\\[-0.8em]
        MinkLoc3Dv2~\cite{komorowski2022minkloc3dv2} & 256 & 49.85 & 49.94 & 82.19 & 71.61 & 51.39 & 50.97 & 80.00 & 71.18 & 75.77 & 84.86 & 67.81 & 79.20 \\
        &&&&&&&&&&&&&\\[-0.8em]
        LoGG3D-Net~\cite{vidanapathirana2022logg3d} & 1024 & 54.03 & \underline{62.40} & 80.37 & 72.47 & 64.26 & 64.05 & 84.54 & 80.26 & 79.84 & 86.55 & 74.67 & 82.66 \\
        &&&&&&&&&&&&&\\[-0.8em]
        ForestLPR\textsuperscript{\textdagger}~\cite{shen2025forestlpr} & 1024 & \textbf{64.15} & \textbf{76.53} & 78.62 & 82.33 & 65.01 & \underline{74.89} & 81.97 & 76.73 & 77.15 & - & \textbf{79.02} & - \\ 
        &&&&&&&&&&&&&\\[-0.8em]
        Ours (C=16, M=16) & 256 & 49.67 & 51.38 & 90.13 & \underline{84.95} & 65.27 & 67.73 & 91.41 & 89.53 & 81.02 & 88.08 & 73.83 & 82.86 \\
        &&&&&&&&&&&&&\\[-0.8em]
        Ours (C=32, M=32) & 1024 & 55.33 & 52.48 & \textbf{92.29} & \textbf{87.62} & \underline{67.56} & 70.56 & \underline{93.33} & \underline{92.31} & \underline{83.94} & \underline{90.05} & 76.12 & \underline{84.67} \\
        &&&&&&&&&&&&&\\[-0.8em]
        Ours (C=128, M=64) & 8192 & \underline{59.33} & 54.02 & \underline{90.49} & \underline{84.95} & \textbf{70.12} & \textbf{77.91} & \textbf{96.38} & \textbf{97.03} & \textbf{86.03} & \textbf{91.20} & \underline{78.83} & \textbf{86.24} \\
        \bottomrule[0.5pt]
    \end{tabular}
    }}
    \label{tab:wildplaces}
\end{table*}

\begin{figure*}[t!]
    \centering
    \includegraphics[width=0.95\textwidth, height=4.6cm]{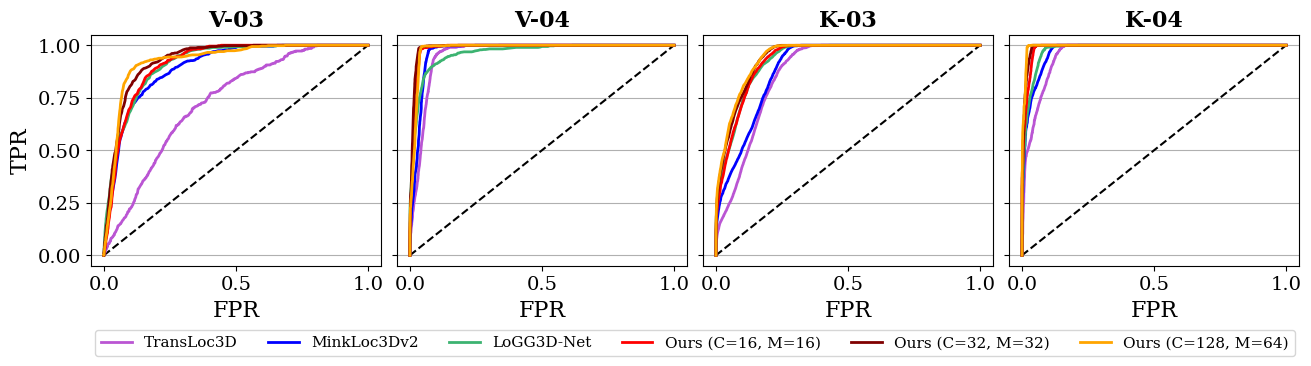}
    \caption{The ROC curves by place recognition models in the Wild-Places intra-sequence evaluations. False positive rate (FPR) and true positive rate (TPR) were measured with the decision threshold in range $[0, 1]$ metrized in the descriptor space.}
    \label{fig:wp_roc}
\end{figure*}

We compared our models to previous methods using NetVLAD (PointNetVLAD~\cite{uy2018pointnetvlad}, TransLoc3D~\cite{xu2021transloc3d}) or GeM (MinkLoc3D~\cite{komorowski2021minkloc3d}, MinkLoc3Dv2~\cite{komorowski2022minkloc3dv2}, CASSPR~\cite{xia2023casspr}, SelFLoc~\cite{qiu2024selfloc}) as an aggregator in the Oxford experiments.
They utilized the pretrained weights published by their authors or were trained from the scratch.
Meanwhile, our methods had two variations; we configured the one have 8192 dimensions of the global descriptor ($C\!\!=\!\!128,\;M\!\!=\!\!64$) to emulate the conventional setup of VLAD and another have 256 dimensions ($C\!\!=\!\!16,\;M\!\!=\!\!16$) for fair comparison to baselines.
Otherwise, they have identical training configurations with $\sigma\!=\!\sqrt{M}$.
The performances of baselines were measured with Recalls at top-1 (R@1) and top-1\% (R@1\%) retrievals.

\Cref{tab:oxford} demonstrates the benefit of our method against previous first-order poolings.
Our method with 8192 dimensions descriptor surpassed recent baselines with advanced backbone architectures, i.e., CASSPR~\cite{xia2023casspr} and SelFLoc~\cite{qiu2024selfloc}, in every measurement.
Meanwhile, our method with 256 dimensions also achieved decent performance gains compared to MinkLoc3Dv2,
implying that the benefit of our method does not rely on the large descriptor size and is manifestable even with the limited capacity of a single cluster.

Moreover, we plotted the Mean Average Precision up to top-10 predictions (MAP@K) in \Cref{fig:curve_oxford}, since Recall@K only indicates whether a correct retrieval exists among top-K predictions but does not reflect how the descriptor space is metrized overall~\cite{musgrave2020metric}.
Our methods showed similar performance tendency to Recall@K;
the one with 8192 dimensions achieved the best, and the one with 256 dimensions also outperformed MinkLoc3Dv2 in the Oxford, U.S., and B.D.
It supplements that our pooling method induces well-separated descriptor space than the existing first-order methods.

\subsection{Experiments on Wild-Places}

The Wild-Places experiments were comprised of two evaluation protocols: the intra-sequence and the inter-sequence.
The intra-sequence evaluates models on the loop-closure detection within a single sequence,
using F1-max and recall@1 measured with the distance in the descriptor space and real-world each.
The inter-sequence assesses models on the global localization with recall@1 and mean rank reciprocal (MRR).
A retrieved scan is considered true when its geographic distance to the query is smaller than 3\;\!m.
While most of the baselines were from \cite{knights2023wildplaces},
we also considered ForestLPR~\cite{shen2025forestlpr}, one of the state-of-the-art methods that utilizes the preprocessing and bird's-eye-view projection with the prior on the natural environment.
Since these baselines have different dimensions of descriptor, e.g., 256 or 1024,
our methods with $\sigma\!=\!M$ have three variations with respect to the dimension as well: 256, 1024, and 8192.

\Cref{tab:wildplaces} shows that our method with the 8192 dimensions achieved the best results among the baselines in the most of inter- and intra-sequence tasks.
Our method with the 1024 dimensions also surpassed LoGG3D-Net and ForestLPR in the V-04 and K-04 intra-sequences and the Venman inter-sequence,
while achieving comparable performances in other evaluations.
Even with the smaller 256 dimensions, our method outperformed MinkLoc3Dv2 and demonstrated comparable performance to the 1024 dimensions.
These results demonstrate that our method is more effective to model complex features from the unstructured environment than relying on the descriptor size or the environment prior.

Furthermore, the ROC curves in \Cref{fig:wp_roc},
which were measured during the intra-sequence evaluations while altering the decision threshold in the descriptor space,
show that our methods formed better curves than other baselines regardless of the descriptor size.
They validate that the better metrizability of the overall descriptor space is inducible with our approach in the natural environment as well.

\subsection{Ablation Studies and Analyses of Our ZCA Whitening}\label{sec:ablation}

\begin{table}[t!]
    \centering
    \caption{Ablation studies in the Oxford and In-house datasets.\\(Top) Ablations on the ZCA whitening.\\(Bottom) Ablations on the backbone architectures.}
    \resizebox{\columnwidth}{!}{{\Large
    \begin{tabular}{l|cccccccc}
        \toprule[1.0pt]
        \multirow{2}{*}{Methods} & \multicolumn{2}{c}{Oxford} & \multicolumn{2}{c}{U.S.} & \multicolumn{2}{c}{R.A.} & \multicolumn{2}{c}{B.D} \\
         & R@1 & R@1\% & R@1 & R@1\% & R@1 & R@1\% & R@1 & R@1\% \\
        \hline
        &&&&&&&&\\[-0.7em]
        \multirow{2}{*}{RBLW~\xmark,\;\;\texttt{SVDPI}~\xmark} & 93.25 & 97.60 & 82.86 & 93.23 & 70.83 & 83.33 & 74.72 & 82.50 \\
                                                & \textcolor{Red}{(-3.38)} & \textcolor{Red}{(-1.27)} & \textcolor{Red}{(-8.18)} & \textcolor{Red}{(-4.13)} & \textcolor{Red}{(-16.13)} & \textcolor{Red}{(-10.56)} & \textcolor{Red}{(-12.83)} & \textcolor{Red}{(-9.26)} \\
        &&&&&&&&\\[-0.7em]
        \multirow{2}{*}{RBLW~\cmark,\;\;\texttt{SVDPI}~\xmark} & \multicolumn{8}{c}{\multirow{2}{*}{--------------------------------------\;\;Diverged\;\;--------------------------------------}}\\\\
        &&&&&&&&\\[-0.7em]
        \multirow{2}{*}{RBLW~\xmark,\;\;\texttt{SVDPI}~\cmark} & 95.82 & 98.57 & 89.18 & 95.75 & 84.95 & 91.47 & 85.12 & 89.94 \\
                                             & \textcolor{Red}{(-0.81)} & \textcolor{Red}{(-0.30)} & \textcolor{Red}{(-1.86)} & \textcolor{Red}{(-1.61)} & \textcolor{Red}{(-2.01)} &\textcolor{Red}{(-2.42)} & \textcolor{Red}{(-2.42)} & \textcolor{Red}{(-1.82)} \\
        \midrule
        \multirow{2}{*}{w/ PointNetVLAD\cite{uy2018pointnetvlad}} & 77.82 & 88.59 & 70.35 & 82.67 & 66.07 & 78.91 & 68.19 & 74.87 \\
                                         & \textcolor{ForestGreen}{(+2.94)} & \textcolor{ForestGreen}{(+0.70)} & \textcolor{ForestGreen}{(+2.26)} & \textcolor{ForestGreen}{(+1.30)} & \textcolor{ForestGreen}{(+2.02)} & \textcolor{ForestGreen}{(+4.25)} & \textcolor{ForestGreen}{(+4.77)} & \textcolor{ForestGreen}{(+4.24)} \\
        &&&&&&&&\\[-0.7em]
        \multirow{2}{*}{w/ MinkLoc3D\cite{komorowski2021minkloc3d}} & 94.88 & 98.26 & 88.27 & 95.75 & 84.47 & 91.46 & 83.89 & 88.78 \\
                                         & \textcolor{ForestGreen}{(+1.12)} & \textcolor{ForestGreen}{(+0.35)} & \textcolor{ForestGreen}{(+2.26)} & \textcolor{ForestGreen}{(+0.71)} & \textcolor{ForestGreen}{(+3.36)} & \textcolor{ForestGreen}{(+0.27)} & \textcolor{ForestGreen}{(+1.23)} & \textcolor{ForestGreen}{(+0.30)} \\
        \bottomrule[1.0pt]
    \end{tabular}
    }}
    \label{tab:ablation}
\end{table}
\begin{figure}[t]
    \centering
    \includegraphics[width=\columnwidth, trim={0.1cm 1.4cm 1.4cm 2.9cm}, clip]{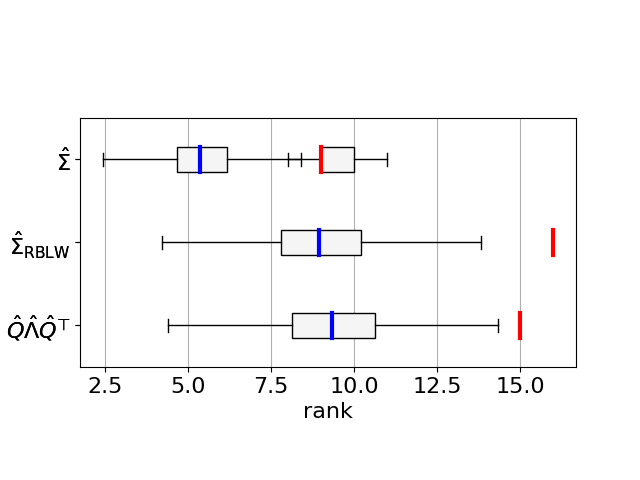}
    \caption{We measured the matrix rank and effective rank~\cite{roy2007effective} of covariances before the whitening ($\hat{\mathbf{\Sigma}}$), after RBLW shrinkage ($\hat{\mathbf{\Sigma}}_\text{RBLW}$), and after the whitening ($\hat{\mathbf{Q}}\hat{\mathbf{\Lambda}}\hat{\mathbf{Q}^{\top}}$) using our default $(C\!=\!16,\;M\!=\!16)$ method. Red and blue lines denote the medians of matrix ranks and effective ranks each.}
    \label{fig:shrink_analysis}
\end{figure}

We conducted quantitative ablation studies in two aspects: (1) the validity of numerical techniques used in our ZCA whitening, and (2) the backbone-agnostic benefit of our approach.
First, we prepared three variants which RBLW is omitted or \texttt{SVDPI} is substituted to a naive eigendecomposition with clipping small eigenvalues.
These variants were trained and evaluated using the identical setups to ours ($C\!\!=\!\!16,\;M\!\!=\!\!16$) in the Oxford.
Second, we also replaced the aggregation layers of PointNetVLAD~\cite{uy2018pointnetvlad} and MinkLoc3D~\cite{komorowski2021minkloc3d} to ours ($C\!\!=\!\!16,\;M\!\!=\!\!16$).
We modified some hyperparameters during their training, e.g., epochs, learning rate, scheduler, and triplet margin,
but their backbone architectures and the loss functions remained unchanged.

The results of ablation studies are in \Cref{tab:ablation} with the performance variance compared to our default method (top) and their original previous methods (bottom) from \Cref{tab:oxford}.
In the ablation of whitening, every alternative method performed worse than our default whitening,
and method with RBLW but without \texttt{SVDPI} even diverged during the training.
Furthermore, our method enhanced the performance of backbones from PointNetVLAD and MinkLoc3D by substituting their first-order aggregators,
supporting its backbone-agnostic benefit to any place recognition models.

We further analyzed the effect of the numerical techniques in our whitening and the cause of the training failure in \Cref{fig:shrink_analysis}.
It shows the change of matirx rank and effective rank~\cite{roy2007effective} of Voronoi cell covariances during the whitening ($\hat{\mathbf{\Sigma}}$, $\hat{\mathbf{\Sigma}}_\text{RBLW}$, $\hat{\mathbf{Q}}\hat{\mathbf{\Lambda}}\hat{\mathbf{Q}^{\top}}$ from \Cref{alg:zca_whitening}),
where the matrix rank was the number of eigenvalues larger than a small threshold ($1\mathrm{e}\!-\!5$) and the effective rank was calculated from the clipped eigenvalues using the identical threshold.

It reveals that RBLW made $\hat{\mathbf{\Sigma}}_\text{RBLW}$ full rank and increased its effective rank as well,
implying that $\hat{\mathbf{\Sigma}}_\text{RBLW}$ embeds more information than $\hat{\mathbf{\Sigma}}$.
However, it also necessitates \texttt{SVDPI} for decomposing the shrunk covariance.
As the discrepancy between the matrix and effective ranks was larger in $\hat{\mathbf{\Sigma}}_\text{RBLW}$,
its repetitive eigenvalues were prone to destabilize the gradient by a naive decomposition than $\hat{\mathbf{\Sigma}}$,
whose gradient by low eigenvalues would be discarded during the low-rank clipping.

\begin{figure}[t]
    \centering
    \includegraphics[width=\columnwidth]{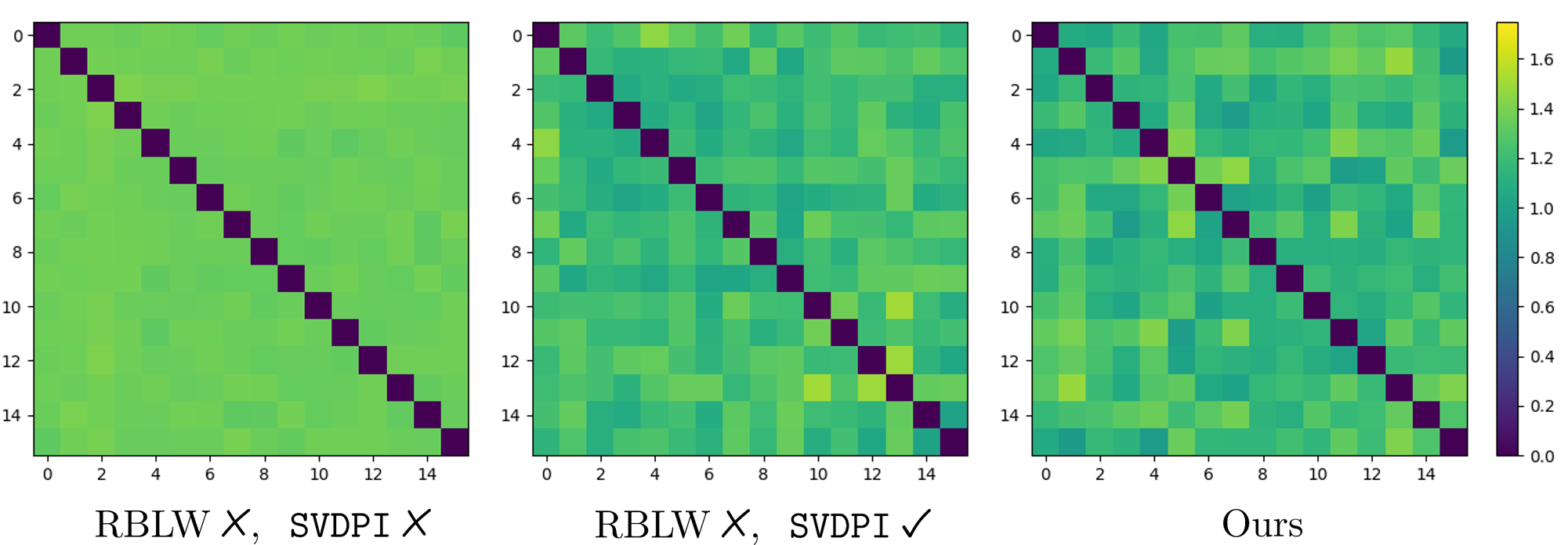}
    \caption{ We measured the empirical Wasserstein-2 distances between the Voronoi cells of descriptors acquired by ablation models in \Cref{tab:ablation}. The brighter color denotes the farther distributional distance between cells.}
    \label{fig:w2_between}
\end{figure}
Moreover, we verified that \texttt{SVDPI} induces more homogeneous Voronoi cell during the training.
\Cref{fig:w2_between} plots the empirical Wasserstein-2 distances between the Voronoi cells on the descriptors by models from ablation studies,
demonstrating that Voronoi cells whitened without \texttt{SVDPI} have larger distributional differences to others.
In summary, both RBLW and \texttt{SVDPI} contribute to the enhanced performance of our ZCA whitening,
and \texttt{SVDPI} especially plays an essential role in stabilizing the descriptor learning.

\section{Conclusion}\label{sec:conclusion}
This paper suggests a novel Voronoi-based second-order pooling method, following the interpretation of connecting NetVLAD and second-order statistics.
Our approach computes outer-products of the local descriptors mapped by separate neural networks,
followed by normalization of the global descriptor using ZCA whitening.
We implement informative and numerically stable ZCA whitening by applying RBLW shrinkage and \texttt{SVDPI}, which enables Mahalanobis metrization across the Voronoi topology.
Extensive quantitative evaluations in the Oxford and Wild-Places demonstrated that the proposed method is performant and forms a well-metrizable descriptor space.
We also verified its backbone agnostic applicability and the validity of RBLW and \texttt{SVDPI} through numerical analyses.
We further expect that our approach will serve as a foundation for future researches in the LPR and VPR as well with its effective metric space regularization and backbone agnostic benefits.

\newpage

\bibliography{ref.bib}

\end{document}